\documentclass{article}

\usepackage{microtype}
\usepackage{graphicx}
\usepackage{subcaption}
\usepackage{booktabs}
\usepackage{multirow}

\usepackage{hyperref}

\usepackage[preprint]{icml2026}

\usepackage{amsmath}
\usepackage{amssymb}
\usepackage{mathtools}
\usepackage{amsthm}

\DeclareMathOperator*{\argmin}{arg\,min}

\usepackage{tikz}
\usetikzlibrary{shapes.geometric, arrows.meta, positioning, calc, fit, backgrounds}

\usepackage[capitalize,noabbrev]{cleveref}

\theoremstyle{plain}

\theoremstyle{definition}

\theoremstyle{remark}

\usepackage[textsize=tiny]{todonotes}

\icmltitlerunning{MoE-Spec: Expert Budgeting for Efficient Speculative Decoding}

\begin{document}

\twocolumn[
  \icmltitle{MoE-Spec: Expert Budgeting for Efficient Speculative Decoding}

  \begin{icmlauthorlist}
    \icmlauthor{Bradley McDanel}{fm,rl}
    \icmlauthor{Steven Li}{rl}
    \icmlauthor{Sruthikesh Surineni}{rl}
    \icmlauthor{Harshit Khaitan}{rl}
  \end{icmlauthorlist}

  \icmlaffiliation{fm}{Franklin and Marshall College}
  \icmlaffiliation{rl}{Meta Reality Labs}
  \icmlcorrespondingauthor{Bradley McDanel}{bmcdanel@fandm.edu}

  \icmlkeywords{Speculative Decoding, Mixture of Experts, Efficient Deep Learning}

  \vskip 0.3in
]

\printAffiliationsAndNotice{}

\begin{abstract}
  Speculative decoding accelerates Large Language Model (LLM) inference by verifying multiple drafted tokens in parallel. However, for Mixture-of-Experts (MoE) models, this parallelism introduces a severe bottleneck: large draft trees activate many unique experts, significantly increasing memory pressure and diminishing speedups from speculative decoding relative to autoregressive decoding. Prior methods reduce speculation depth when MoE verification becomes expensive. We propose MoE-Spec, a training-free verification-time expert budgeting method that decouples speculation depth from memory cost by enforcing a fixed expert capacity limit at each layer, loading only the experts that contribute most to verification and dropping the long tail of rarely used experts that drive bandwidth overhead. Experiments across multiple model scales and datasets show that this method yields 10--30\% higher throughput than state-of-the-art speculative decoding baselines (EAGLE-3) at comparable quality, with flexibility to trade accuracy for further latency reductions through tighter budgets.
\end{abstract}

\section{Introduction}
\label{sec:intro}

\begin{figure}[t]
  \centering
  \includegraphics[width=\columnwidth]{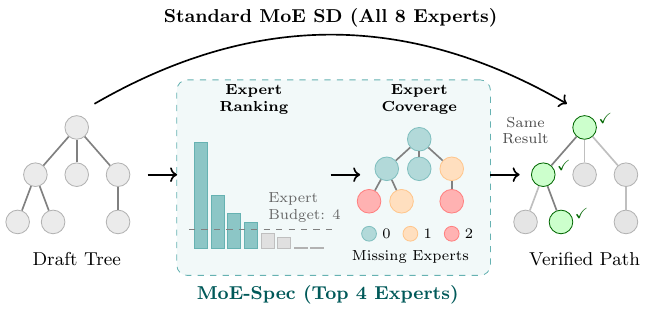}
  \caption{Overview of MoE-Spec. Standard MoE speculative decoding loads all N experts (N = 8) activated across the draft tree (top path). MoE-Spec ranks experts by aggregate routing probability and enforces a budget $B$ (here $B=4$), loading only top-scoring experts (bottom path). Tokens are color-coded by the number of missing experts from their natural routing. In this example, both paths accept the same three tokens despite MoE-Spec loading half as many experts.}
  \label{fig:overview}
\end{figure}

Mixture-of-Experts (MoE) models have become the dominant architecture for
state-of-the-art language models, powering systems such as GPT-4~\citep{achiam2023gpt},
DeepSeek-R1~\citep{guo2025deepseek}, and Llama-4~\citep{meta2024llama4}. MoEs reduce
inference cost through sparse activation: each token activates only $k$ of the $N$
total experts per layer, so memory bandwidth scales with active rather than total
parameters. Speculative decoding accelerates LLM inference by verifying multiple drafted
tokens in parallel~\citep{speculative-decoding, chen2023accelerating}. For dense models,
this parallelism is essentially free: all parameters are loaded once per forward pass,
so verification cost is constant regardless of draft size.

\begin{figure*}[t]
  \centering
  \begin{subfigure}[b]{0.42\textwidth}
    \includegraphics[width=\textwidth]{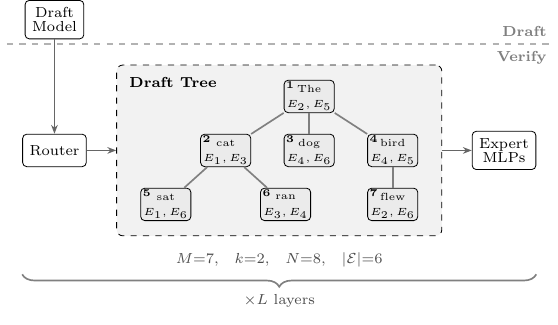}
    \caption{MoE verification}
    \label{fig:motivation-flow}
  \end{subfigure}
  \hfill
  \begin{subfigure}[b]{0.27\textwidth}
    \includegraphics[width=\textwidth]{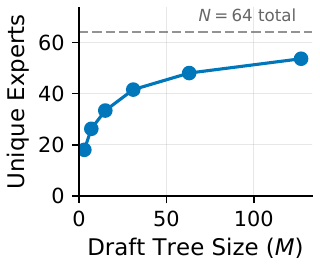}
    \caption{Expert explosion}
    \label{fig:motivation-explosion}
  \end{subfigure}
  \hfill
  \begin{subfigure}[b]{0.27\textwidth}
    \includegraphics[width=\textwidth]{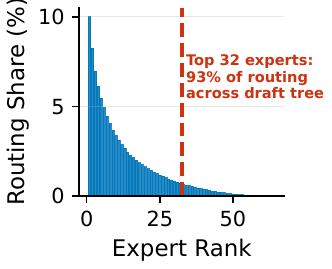}
    \caption{Heavy-tailed routing}
    \label{fig:motivation-zipf}
  \end{subfigure}
  \caption{Motivation for expert budgeting. (a) During verification, each token routes to $k$ experts; the target model must load all unique experts $\mathcal{E}$ across the draft tree. (b) As tree size $M$ grows, unique experts per layer approaches $N$, negating sparse activation benefits. (c) Routing probabilities are heavy-tailed: the top 32 of 64 experts capture 93\% of routing weight for a tree size of 63.}
  \label{fig:motivation}
\end{figure*}

For MoE models, verification cost is no longer constant: it scales with the diversity of expert activations across the draft tree. Each token routes independently to its own subset of experts, and the target model must load the \emph{union} of all selected experts to verify the tree (\Cref{fig:motivation-flow}). As draft size grows, so does the number of unique experts: on OLMoE-1B-7B, a 127-token tree activates 54 of 64 experts per layer, approaching the full model (\Cref{fig:motivation-explosion}). Verification can thus become more expensive than autoregressive decoding, negating both the benefits of sparse activation and speculation.

However, not all experts in the union contribute equally to verification. Expert activations across a draft tree follow a heavy-tailed distribution (\Cref{fig:motivation-zipf}): on OLMoE, the top 32 of 64 experts capture 93\% of aggregate routing probability. The remaining experts incur bandwidth cost but contribute little to output quality. This asymmetry suggests that verification cost can be reduced without proportional quality loss by enforcing an expert budget: loading only the highest-scoring experts and dropping the long tail.

We propose MoE-Spec, a method that enforces a strict expert budget during verification (\Cref{fig:overview}). Given a draft tree, we aggregate routing probabilities across all tokens and select the top-$B$ experts to form a shortlist. Tokens whose natural routing falls outside the shortlist are handled by truncation (omitting missing experts) or substitution (replacing them with available alternatives). This decouples verification cost from draft tree complexity: regardless of tree size, we load at most $B$ experts per layer. We explore three ranking strategies: static (fixed ordering from calibration), router-based (aggregating routing probabilities per tree), and an Oracle upper bound that greedily minimizes reconstruction error.

We evaluate MoE-Spec on three MoE architectures (OLMoE-1B-7B, Qwen3-30B-A3B, Mixtral-8x7B) across mathematical reasoning, code generation, and summarization tasks. Compared to EAGLE-3~\citep{li2025eagle3}, MoE-Spec achieves 10--30\% higher throughput at comparable quality, with flexibility to trade accuracy for further speedup through tighter budgets. Our main contributions are:
\begin{itemize}
    \item We show that expert activations during MoE verification are heavy-tailed, with a small subset capturing most routing weight across the draft tree.
    \item We propose MoE-Spec, a simple yet effective method that enforces an expert budget during verification by selecting a budgeted subset of these top-scoring experts. Unlike prior work that requires architectural changes or additional training, our approach is training-free and integrates directly into existing speculative decoding pipelines.    \item We demonstrate throughput improvements over EAGLE-3 across three MoE architectures spanning different scales and expert configurations.
    \item Our Oracle analysis indicates that improved ranking could reduce expert requirements by 25\% over our router-based heuristic, motivating future work on better selection methods.
\end{itemize}

\section{Background}
\label{sec:background}

\subsection{Mixture-of-Experts}
\label{sec:background:moe}

Mixture-of-Experts (MoE) models achieve parameter efficiency by activating only a subset of model weights for each token. In an MoE layer, the standard feed-forward network is replaced with $N$ parallel expert networks $\{E_1, \ldots, E_N\}$ and a learned router that selects $k$ experts to process each token (where $k \ll N$).

Given a token representation $\mathbf{h} \in \mathbb{R}^d$, the router computes a probability distribution over experts via a linear projection followed by softmax:\footnote{Some architectures, such as DeepSeek-V3, use sigmoid-based routing instead. Our method applies to both, though we focus on softmax-routed models in experiments.}
\begin{equation}
  g_i(\mathbf{h}) = \frac{\exp(\mathbf{w}_i^\top \mathbf{h})}{\sum_{j=1}^{N} \exp(\mathbf{w}_j^\top \mathbf{h})},
  \label{eq:routing-prob}
\end{equation}
where $\mathbf{w}_i \in \mathbb{R}^d$ is the routing weight vector for expert $i$. Let $\mathcal{T}_k(\mathbf{h}) \subset \{1, \ldots, N\}$ denote the indices of the top-$k$ experts by routing probability. The MoE layer output is a weighted combination of the selected experts:
\begin{equation}
  \text{MoE}(\mathbf{h}) = \sum_{i \in \mathcal{T}_k(\mathbf{h})} \bar{g}_i(\mathbf{h}) \cdot E_i(\mathbf{h}),
  \label{eq:moe-output}
\end{equation}
where $\bar{g}_i(\mathbf{h})$ denotes the mixing weights. These are either the raw routing probabilities $g_i(\mathbf{h})$ or renormalized values $g_i(\mathbf{h}) / \sum_{j \in \mathcal{T}_k(\mathbf{h})} g_j(\mathbf{h})$, depending on the architecture (e.g., OLMoE renormalizes; Qwen3 does not).

This sparse activation pattern means that inference cost scales with the number of \emph{active} parameters rather than total parameters. For a single token, only $k$ experts (typically $k \in \{2, 8\}$) are loaded from memory. However, when processing multiple tokens jointly, the model must load the \emph{union} of experts selected across all tokens, which can approach the full set of $N$ experts as token count grows.

\subsection{Speculative Decoding with Draft Trees}
\label{sec:background:spec}

Speculative decoding accelerates autoregressive generation by drafting multiple candidate tokens and verifying them in parallel~\citep{speculative-decoding, chen2023accelerating}. A lightweight \emph{draft model} proposes candidate continuations, and the full \emph{target model} verifies these candidates in a single forward pass. Accepted tokens are emitted immediately; rejected tokens trigger resampling from a corrected distribution. This process is provably lossless for greedy decoding as the output distribution matches standard autoregressive decoding.

Modern methods such as EAGLE~\citep{li2024eagle, li2024eagle2} extend this to \emph{tree-structured} drafts, where multiple candidate tokens are proposed at each position. The target model verifies all candidates simultaneously using tree attention, accepting a path of tokens consistent with the target distribution.

For dense models, verification cost is effectively constant regardless of draft tree size: all model parameters are loaded once per forward pass, and the additional compute for extra tokens is negligible in the memory-bound regime. This makes aggressive speculation attractive---larger trees increase the expected number of accepted tokens per iteration. For MoE models, however, this assumption breaks down; we analyze this problem in \Cref{sec:method:problem}.

\section{Method}
\label{sec:method}

\subsection{Verification Overhead in Tree Decoding}
\label{sec:method:problem}

As described in \Cref{sec:intro}, MoE verification cost scales with expert diversity across the draft tree. We formalize this as follows. Let $\mathcal{D} = \{\mathbf{h}_1, \ldots, \mathbf{h}_M\}$ denote a draft tree of $M$ token representations. To verify the tree, the target model must load the union of all selected experts:
\begin{equation}
  \mathcal{E}(\mathcal{D}) = \bigcup_{t=1}^{M} \mathcal{T}_k(\mathbf{h}_t).
  \label{eq:expert-union}
\end{equation}
As \Cref{fig:motivation-explosion} shows, $|\mathcal{E}|$ approaches $N$ for large trees, negating sparse activation benefits. However, expert routing is heavy-tailed (\Cref{fig:motivation-zipf}): by enforcing a budget $B < |\mathcal{E}|$ and loading only the highest-scoring experts, we reduce bandwidth while preserving verification quality.

\subsection{Expert Importance Ranking}
\label{sec:method:ranking}

Given an expert budget $B < N$, we select a shortlist $\mathcal{S}$ of $B$ experts to load during verification. We consider three ranking strategies.

\paragraph{Static Ranking.}
The simplest approach precomputes a fixed ordering from calibration data by counting selection frequency:
\begin{equation}
  c_i = \sum_{\mathbf{h} \in \text{calib}} \mathbf{1}[i \in \mathcal{T}_k(\mathbf{h})].
  \label{eq:static-count}
\end{equation}
The shortlist consists of the $B$ experts with highest $c_i$. This incurs zero runtime overhead and enables permanent model pruning---deploying only 32 of 64 experts halves memory requirements. At moderate budgets ($B \geq N/2$), static ranking performs comparably to adaptive methods on most tasks (\Cref{appendix:coverage}). The gap widens at aggressive budgets, where input-dependent routing becomes more valuable.

\paragraph{Router-Based Ranking.}
Our primary method aggregates routing probabilities across the draft tree:
\begin{equation}
  s_i = \sum_{t=1}^{M} g_i(\mathbf{h}_t).
  \label{eq:router-score}
\end{equation}
The shortlist $\mathcal{S}$ consists of the $B$ experts with highest $s_i$. This requires no computation beyond what the router already produces, and adapts to each tree's routing pattern.

\paragraph{Oracle Ranking.}
As an upper bound, we consider an oracle with access to the target model's routing decisions and expert outputs. The oracle greedily selects experts to minimize reconstruction error against the unpruned output. Let $O^* = \sum_{i \in \mathcal{T}_k(\mathbf{h}_t)} g_i(\mathbf{h}_t) \cdot E_i(\mathbf{h}_t)$ denote the gold output. Starting from $\mathcal{S} = \emptyset$, the oracle iteratively adds the expert $i^*$ that minimizes:
\begin{equation}
  i^* = \argmin_{i \notin \mathcal{S}} \sum_{t=1}^{M} \left\| O^*_t - \sum_{j \in \mathcal{S} \cup \{i\}} g_j(\mathbf{h}_t) \cdot E_j(\mathbf{h}_t) \right\|_2^2.
  \label{eq:oracle-greedy}
\end{equation}
This requires running all experts and is impractical for speedup, but provides a ceiling on achievable quality.

\subsection{Expert Coverage}
\label{sec:method:coverage}

Given shortlist $\mathcal{S}$, we must handle tokens whose natural routing includes experts outside $\mathcal{S}$. We consider two policies.

\paragraph{Truncation.}
Each token uses its natural top-$k$ routing, but experts outside $\mathcal{S}$ contribute zero:
\begin{equation}
  \text{MoE}_{\mathcal{S}}^{\text{trunc}}(\mathbf{h}_t) = \sum_{i \in \mathcal{T}_k(\mathbf{h}_t) \cap \mathcal{S}} \bar{g}_i(\mathbf{h}_t) \cdot E_i(\mathbf{h}_t).
  \label{eq:truncation}
\end{equation}
Probabilities $\bar{g}_i$ are computed over the original top-$k$ set (not re-computed for the reduced set). Tokens whose natural routing overlaps partially with $\mathcal{S}$ receive fewer than $k$ experts. In the extreme case where $\mathcal{T}_k(\mathbf{h}_t) \cap \mathcal{S} = \emptyset$, the MoE output is zero and only the residual connection passes through. This effectively skips the MoE computation for that token at that layer.

\paragraph{Substitution.}
Top-$k$ selection is constrained to $\mathcal{S}$. Let $\mathcal{T}_k^{\mathcal{S}}(\mathbf{h}_t)$ denote the top-$k$ experts within $\mathcal{S}$, ranked by $g_i(\mathbf{h}_t)$:
\begin{equation}
  \text{MoE}_{\mathcal{S}}^{\text{sub}}(\mathbf{h}_t) = \sum_{i \in \mathcal{T}_k^{\mathcal{S}}(\mathbf{h}_t)} \bar{g}_i^{\mathcal{S}}(\mathbf{h}_t) \cdot E_i(\mathbf{h}_t),
  \label{eq:substitution}
\end{equation}
where $\bar{g}_i^{\mathcal{S}}$ are the routing probabilities over the constrained selection. If a token's natural top-$k$ includes experts outside $\mathcal{S}$, those experts are replaced by the highest-scoring available alternatives within the shortlist. Every token receives exactly $k$ experts, though they may differ from natural routing. We compare both policies empirically in \Cref{sec:exp:ablations}.

\section{Experiments}
\label{sec:experiments}

\subsection{Experimental Setup}
\label{sec:exp:setup}

\paragraph{Models.}
We evaluate on three MoE architectures spanning different scales and expert configurations: OLMoE-1B-7B~\citep{muennighoff2024olmoe}, with 64 experts and $k=8$ active per token (7B total, 1B active parameters); Qwen3-30B-A3B, with 128 experts and $k=8$ (30B total, 3B active); and Mixtral-8x7B~\citep{jiang2024mixtral}, with 8 experts and $k=2$ (47B total, 13B active). For speculative decoding, we use EAGLE-3~\citep{li2025eagle3} draft models for OLMoE and Qwen3, and the original EAGLE~\citep{li2024eagle} for Mixtral.

\paragraph{Benchmarks.}
We evaluate on five benchmarks spanning mathematical reasoning, code generation, and summarization. GSM8K and MATH500 test multi-step arithmetic and competition-level math problems respectively, measured by accuracy. HumanEval and MBPP evaluate code generation via pass@1 on function synthesis tasks. CNN/DailyMail tests abstractive summarization, measured by ROUGE-L F1.

\paragraph{Baselines.}
We compare against two baselines: autoregressive decoding (AR), which generates tokens sequentially without speculation, and EAGLE, the current state-of-the-art for tree-structured speculative decoding. We use EAGLE-3 for OLMoE and Qwen3, and EAGLE-1 for Mixtral where EAGLE-3 draft models are unavailable. All draft model adapters are from the official EAGLE repository.

\paragraph{Implementation.}
We implement MoE-Spec by extending the EAGLE-3 codebase\footnote{\url{https://github.com/SafeAILab/EAGLE}} to add expert budgeting during the verification phase (\Cref{appendix:algorithm}). We use the open-source EAGLE-3 implementation without modification to tree construction or verification; our only change is batched MoE computation for efficiency (\Cref{appendix:batched}). All main results use router-based expert selection with the substitution policy; ablations over selection methods and coverage policies appear in \Cref{sec:exp:ablations}. Following~\citet{li2025eagle3}, we use draft trees of 63 tokens and evaluate at both temperature 0 (greedy decoding) and temperature 1 (sampling).\footnote{Lossless speculative decoding requires rejection sampling with draft probability $q(x)$~\citep{speculative-decoding}. For tree-based methods, computing $q(x)$ requires tracking probabilities across all paths. The EAGLE implementation uses $q(x)=1$, which causes the distribution shift we observe at $T{=}1$. MoE-Spec inherits this, so comparisons remain valid.} Full hyperparameters and model configurations are in \Cref{appendix:hyperparameters}.

\subsection{Main Results}
\label{sec:exp:main}

MoE-Spec improves speedup over EAGLE in 27 of 30 model-benchmark configurations while maintaining comparable quality. \Cref{tab:main-results} compares autoregressive decoding (AR), EAGLE, and MoE-Spec across three MoE architectures and five benchmarks at both $T=0$ and $T=1$. For MoE-Spec, we select the expert budget that maximizes throughput while keeping quality within one standard deviation of EAGLE (measured from $T=1$ runs with 5 seeds). As a concrete example, on Mixtral MBPP at $T=0$, MoE-Spec achieves 2.3$\times$ speedup compared to EAGLE's 1.7$\times$, a 35\% relative improvement at identical task accuracy.

\begin{table*}[t]
\centering
\caption{Speedup over autoregressive decoding, mean acceptance length ($\tau$), and task metrics at T=0 and T=1. MoE-Spec uses substitution policy with the expert budget maximizing throughput while maintaining quality comparable to EAGLE-3 (within one standard deviation, measured from T=1 runs with 5 seeds). T=1 results report mean over 5 random seeds.}
\label{tab:main-results}
\small
\begin{tabular}{@{}ll@{\qquad}c@{\;\;}c@{\;\;}c@{\qquad}c@{\;\;}c@{\;\;}c@{\qquad}c@{\;\;}c@{\;\;}c@{\qquad}c@{\;\;}c@{\;\;}c@{\qquad}c@{\;\;}c@{\;\;}c@{}}
\toprule
 &  & \multicolumn{3}{c}{GSM8K} & \multicolumn{3}{c}{MATH500} & \multicolumn{3}{c}{HumanEval} & \multicolumn{3}{c}{MBPP} & \multicolumn{3}{c}{CNN/DM} \\
\cmidrule(lr){3-5} \cmidrule(lr){6-8} \cmidrule(lr){9-11} \cmidrule(lr){12-14} \cmidrule(lr){15-17}
Model & Method & Spd & Acc & $\tau$ & Spd & Acc & $\tau$ & Spd & P@1 & $\tau$ & Spd & P@1 & $\tau$ & Spd & R-L & $\tau$ \\
\midrule
\multicolumn{17}{l}{\textbf{Temperature = 0}} \\
\midrule
Mixtral-8x7B & AR & 1.0$\times$ & 39 & -- & 1.0$\times$ & 31 & -- & 1.0$\times$ & 22 & -- & 1.0$\times$ & 56 & -- & 1.0$\times$ & 18 & -- \\
 & EAGLE & 1.5$\times$ & 46 & 2.8 & 1.4$\times$ & 31 & 2.5 & 1.6$\times$ & 20 & 3.2 & 1.7$\times$ & 45 & 3.2 & 1.3$\times$ & 18 & 2.3 \\
 & MoE-Spec & \textbf{1.8$\times$} & 44 & 2.9 & \textbf{2.0$\times$} & \textbf{34} & 2.5 & \textbf{1.9$\times$} & \textbf{27} & 3.1 & \textbf{2.3$\times$} & 45 & 3.1 & \textbf{1.5$\times$} & 18 & 2.2 \\
\midrule
OLMoE-1B-7B & AR & 1.0$\times$ & 74 & -- & 1.0$\times$ & 28 & -- & 1.0$\times$ & 30 & -- & 1.0$\times$ & 45 & -- & 1.0$\times$ & 18 & -- \\
 & EAGLE-3 & 1.6$\times$ & 69 & 3.0 & 2.1$\times$ & 26 & 3.6 & 2.3$\times$ & 30 & 4.2 & 2.2$\times$ & 45 & 4.0 & 1.2$\times$ & 17 & 2.0 \\
 & MoE-Spec & \textbf{1.7$\times$} & 66 & 3.0 & \textbf{2.5$\times$} & \textbf{28} & 3.5 & \textbf{2.4$\times$} & 30 & 4.2 & 2.2$\times$ & \textbf{46} & 3.8 & \textbf{1.3$\times$} & \textbf{18} & 2.0 \\
\midrule
Qwen3-30B-A3B & AR & 1.0$\times$ & 86 & -- & 1.0$\times$ & 82 & -- & 1.0$\times$ & 82 & -- & 1.0$\times$ & 90 & -- & 1.0$\times$ & 16 & -- \\
 & EAGLE-3 & 1.9$\times$ & 85 & 3.2 & 1.8$\times$ & 81 & 3.0 & 1.8$\times$ & 82 & 3.1 & 1.7$\times$ & 89 & 2.9 & 1.5$\times$ & 17 & 2.1 \\
 & MoE-Spec & \textbf{2.4$\times$} & 84 & 3.2 & \textbf{2.2$\times$} & 81 & 2.9 & \textbf{2.2$\times$} & 78 & 3.1 & 1.7$\times$ & 89 & 2.9 & \textbf{1.9$\times$} & 16 & 2.0 \\
\midrule
\multicolumn{17}{l}{\textbf{Temperature = 1}} \\
\midrule
Mixtral-8x7B & AR & 1.0$\times$ & 44 & -- & 1.0$\times$ & 30 & -- & 1.0$\times$ & 21 & -- & 1.0$\times$ & 57 & -- & 1.0$\times$ & 18 & -- \\
 & EAGLE & 1.5$\times$ & 42 & 2.8 & 1.4$\times$ & 30 & 2.5 & 1.6$\times$ & 21 & 3.1 & 1.6$\times$ & 48 & 3.1 & 1.2$\times$ & 18 & 2.2 \\
 & MoE-Spec & \textbf{2.1$\times$} & 40 & 2.8 & \textbf{1.6$\times$} & \textbf{31} & 2.5 & \textbf{1.9$\times$} & 20 & 3.1 & \textbf{1.8$\times$} & \textbf{50} & 3.1 & \textbf{1.8$\times$} & 18 & 2.1 \\
\midrule
OLMoE-1B-7B & AR & 1.0$\times$ & 60 & -- & 1.0$\times$ & 26 & -- & 1.0$\times$ & 29 & -- & 1.0$\times$ & 41 & -- & 1.0$\times$ & 15 & -- \\
 & EAGLE-3 & 1.7$\times$ & 49 & 3.2 & 1.9$\times$ & 22 & 3.4 & 2.0$\times$ & 26 & 3.6 & 2.0$\times$ & 40 & 3.5 & 1.0$\times$ & 16 & 1.4 \\
 & MoE-Spec & \textbf{1.9$\times$} & \textbf{53} & 3.1 & \textbf{2.1$\times$} & \textbf{26} & 3.2 & 2.0$\times$ & \textbf{29} & 3.8 & 2.0$\times$ & \textbf{42} & 3.5 & 1.0$\times$ & 15 & 1.4 \\
\midrule
Qwen3-30B-A3B & AR & 1.0$\times$ & 84 & -- & 1.0$\times$ & 79 & -- & 1.0$\times$ & 81 & -- & 1.0$\times$ & 88 & -- & 1.0$\times$ & 16 & -- \\
 & EAGLE-3 & 1.8$\times$ & 87 & 3.2 & 1.8$\times$ & 79 & 2.9 & 1.8$\times$ & 80 & 3.0 & 1.7$\times$ & 89 & 2.9 & 1.5$\times$ & 16 & 2.0 \\
 & MoE-Spec & \textbf{2.1$\times$} & 84 & 3.2 & \textbf{2.3$\times$} & 79 & 2.8 & \textbf{1.9$\times$} & 78 & 3.0 & 1.7$\times$ & 89 & 2.9 & \textbf{1.7$\times$} & 16 & 2.0 \\
\bottomrule
\end{tabular}
\end{table*}

\begin{figure}
  \centering
  \includegraphics[width=\columnwidth]{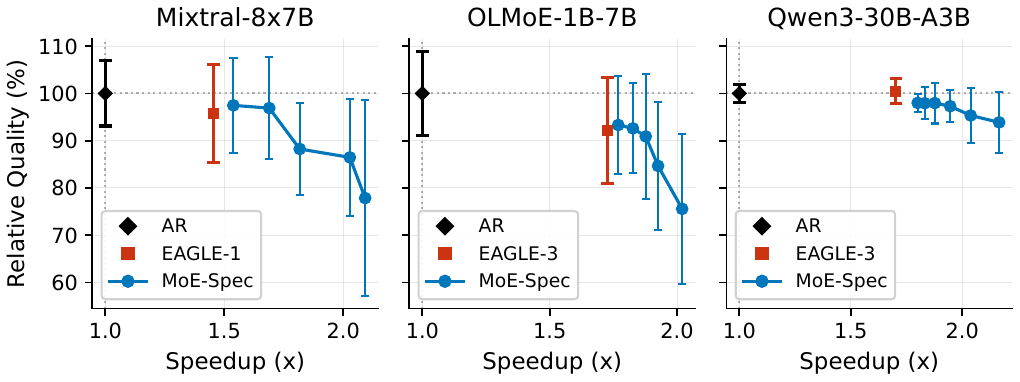}
  \caption{Quality-speedup tradeoff at $T=1$, averaged across five benchmarks. For each benchmark, quality and speedup are normalized relative to AR (100\%), then averaged. Error bars combine cross-benchmark variance with seed-to-seed variance from 5 runs. Individual per-benchmark curves appear in \Cref{appendix:pareto}.}
  \label{fig:pareto-summary}
\end{figure}

Larger models show greater relative improvements from expert budgeting. Mixtral-8x7B (14B active parameters) averages 27\% relative speedup gain over EAGLE across both temperatures. Qwen3-30B-A3B (3B active) averages 16\%, and OLMoE-1B-7B (1B active) averages 6\%. MoE-Spec achieves similar acceptance lengths to EAGLE despite loading fewer experts, with only a 1.4\% reduction on average across all configurations. Since acceptance rates remain nearly unchanged, the speedup gains come entirely from reduced verification cost per step.

Given this robustness, the expert budget becomes a practical control surface for balancing quality against latency. \Cref{fig:pareto-summary} shows the tradeoff at $T=1$: for each benchmark, we normalize quality and speedup relative to AR, then average across all five benchmarks. Unlike EAGLE, which provides a single operating point, MoE-Spec traces a continuous Pareto curve from conservative (high budget, near-EAGLE quality) to aggressive (low budget, higher speedup). This enables applications to select operating points tailored to their quality requirements.

The quality degradation profile varies across models. Qwen3-30B-A3B shows the most favorable tradeoff, maintaining approximately 95\% quality even at 2.0$\times$ speedup with tight error bars. Mixtral-8x7B exhibits moderate degradation, with quality dropping to approximately 78\% at maximum speedup. OLMoE-1B-7B shows the steepest decline and highest variance at aggressive speedup targets, consistent with smaller models being more sensitive to approximations. \Cref{tab:main-results} reports one point on each curve: the budget that maximizes speedup while staying within quality tolerance. Individual per-benchmark curves appear in \Cref{appendix:pareto}.

\subsection{Speedup Analysis}
\label{sec:exp:speedup}

\begin{figure*}[t]
  \centering
  \begin{minipage}[b][0pt][b]{0pt}\phantomsubcaption\label{fig:mechanism-activation}\end{minipage}%
  \begin{minipage}[b][0pt][b]{0pt}\phantomsubcaption\label{fig:mechanism-speedup}\end{minipage}%
  \begin{minipage}[b][0pt][b]{0pt}\phantomsubcaption\label{fig:mechanism-experts}\end{minipage}%
  \includegraphics[width=\textwidth]{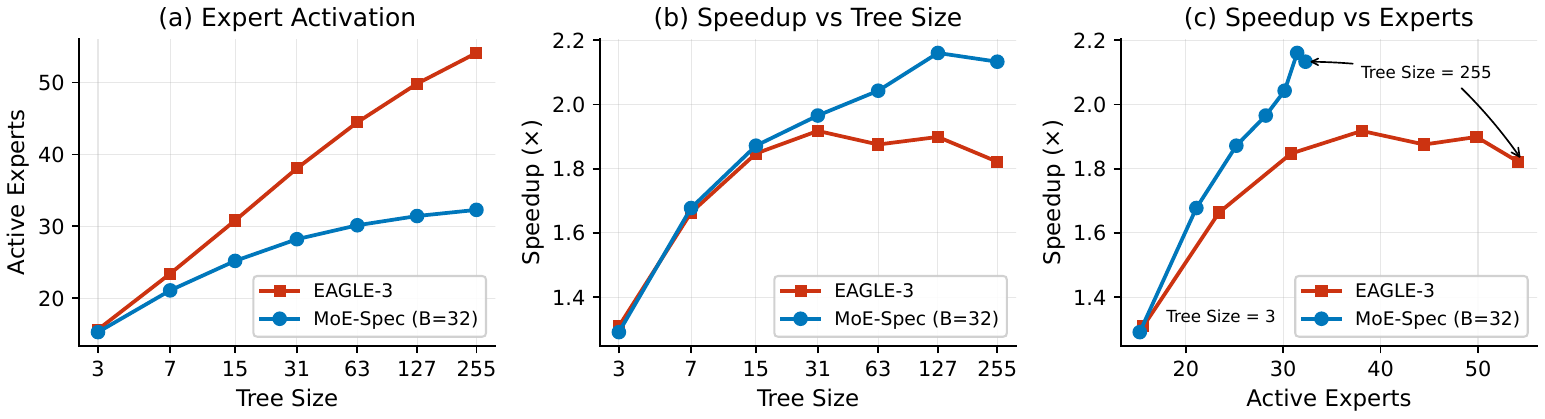}
  \caption{Expert activation and speedup on OLMoE-1B-7B. (\subref{fig:mechanism-activation}) Unique experts activated during verification as tree size increases. EAGLE-3 loads over 50 of 64 experts at large trees; MoE-Spec with $B=32$ saturates at the budget. (\subref{fig:mechanism-speedup}) Speedup relative to AR. EAGLE-3 peaks at tree size 31 then declines as expert loading dominates; MoE-Spec continues improving at larger trees. (\subref{fig:mechanism-experts}) Speedup vs.\ active experts. At tree size 255, EAGLE-3 loads 54 experts for 1.85$\times$ speedup; MoE-Spec loads 32 experts for 2.1$\times$ speedup.}
  \label{fig:mechanism}
\end{figure*}

The speedup gains in \cref{tab:main-results} arise because MoE-Spec loads at most $B$ experts per layer while EAGLE-3 must load all experts activated across the draft tree, even those used by only a minority of tokens (\cref{fig:motivation-zipf}). \Cref{fig:mechanism} illustrates this effect on OLMoE-1B-7B at a fixed expert budget $B=32$ as a function of tree size. For this experiment, we aggregate across all 5 datasets outlined in \cref{sec:exp:setup}. \Cref{fig:mechanism-activation} shows that the average active number of experts per draft tree is significantly smaller for MoE-Spec compared to EAGLE-3 even at moderate tree sizes. For example, at tree size of 31, MoE-Spec uses on average 29 experts compared to 39 for EAGLE-3. At larger tree sizes, this gap widens: MoE-Spec saturates at 32 experts while EAGLE-3 approaches using all 64 experts per tree.

The speedup for EAGLE peaks at tree size 31 then declines as straggler experts dominate verification cost, while MoE-Spec continues improving throughout (\cref{fig:mechanism-speedup}). Both methods achieve similar gains at small tree sizes where expert counts remain low. Beyond tree size 15, EAGLE peaks at 1.95$\times$ then declines to 1.85$\times$ at tree size 255. The memory bandwidth cost of loading straggler experts, those activated by only a few tokens in the tree, outpaces the benefit from longer accepted sequences. MoE-Spec avoids this overhead and continues improving throughout, reaching 2.1$\times$ at tree size 255.

The relationship between speedup and active experts reveals a saturation point around 32 experts, beyond which additional loading yields diminishing returns. \Cref{fig:mechanism-experts} shows this by plotting speedup against active experts directly. EAGLE operates past this saturation point at large tree sizes, reaching 54 experts at tree size 255, while MoE-Spec stays below it by construction. The experts beyond the saturation point are stragglers from the tail of the routing distribution: they serve only a handful of tokens in each draft tree but incur the same memory bandwidth cost as the heavily-used experts. Expert budgeting excludes them from verification, keeping cost bounded while preserving the high-value experts that dominate routing probability (\cref{fig:motivation-zipf}). The selection overhead (computing routing probabilities and sorting to form the shortlist) consumes only 2--3\% of total generation time, far smaller than the verification savings from reduced expert loading.

The robustness to aggressive budgets follows from the routing distribution: at $B=32$, the excluded experts collectively receive only 7\% of routing weight across the draft tree. The router's own probability assignments identify which experts matter, and budgeting simply enforces that judgment. For tokens whose preferred experts fall outside the shortlist, substitution provides alternative experts while truncation relies on the residual connection to preserve the hidden state.

\subsection{Ablation Studies}
\label{sec:exp:ablations}

\begin{figure}[t]
  \centering
  \begin{minipage}[b][0pt][b]{0pt}\phantomsubcaption\label{fig:ablation-selection}\end{minipage}%
  \begin{minipage}[b][0pt][b]{0pt}\phantomsubcaption\label{fig:ablation-policy}\end{minipage}%
  \includegraphics[width=\columnwidth]{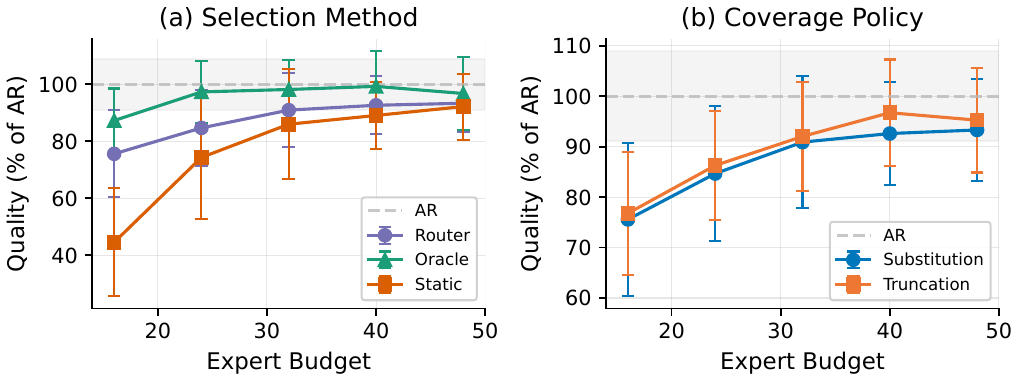}
  \caption{Design ablations on OLMoE-1B-7B at $T=1$, normalized to autoregressive (AR) baseline. (a) Selection methods: Static ranking fails at low budgets while Router-based ranking tracks the Oracle upper bound. (b) Coverage policies: Truncation and Substitution perform comparably across budgets.}
  \label{fig:ablation-olmoe}
\end{figure}

We evaluate the impact of the expert selection strategy on generation quality across varying budgets. \Cref{fig:ablation-selection} compares three ranking methods using the Substitution coverage policy: Router-based (aggregating routing probabilities across the draft tree), Static (fixed ordering from calibration data), and Oracle (greedy selection using target model outputs). Static ranking fails at low budgets ($B \leq 16$), achieving near-zero accuracy on GSM8K and MATH500 because a fixed expert set cannot adapt to input variability. Router-based ranking remains robust across the budget range, tracking the Oracle upper bound even at tight budgets.

The Oracle method achieves near-AR quality even at $B=24$ (37.5\% of experts), suggesting room for improved selection strategies. While the Oracle is impractical for acceleration (it requires executing the full model to obtain routing decisions), better selection methods could close the remaining gap between Router and Oracle. Router-based ranking offers the best practical tradeoff: it incurs no additional model execution cost while outperforming Static ranking by a wide margin at low budgets, and converging to within 5--10\% of AR at moderate budgets ($B \geq 32$). We use Router-based ranking as the default.

\Cref{fig:ablation-policy} examines how to handle tokens whose preferred experts fall outside the selected budget, using Router-based selection. Truncation zeros out missing experts while preserving natural routing weights; Substitution replaces missing experts with the highest-scoring available alternatives within the shortlist. Both policies perform comparably across all benchmarks and expert budgets. We select Substitution as the default to maintain a constant number of active experts ($k$) per token.

\subsection{Reconstruction Error Analysis}
\label{sec:exp:reconstruction}

Router-based ranking tracks but does not match Oracle across the budget range. To understand this gap, we measure reconstruction error: the normalized squared distance between the budgeted MoE output $O_{\mathcal{S}}$ and the full output $O^*$ from \Cref{eq:moe-output}. We compute this at each MoE layer during Oracle selection, where all expert outputs are available, and average across layers and datasets.

\begin{figure}
  \centering
  \includegraphics[width=0.8\columnwidth]{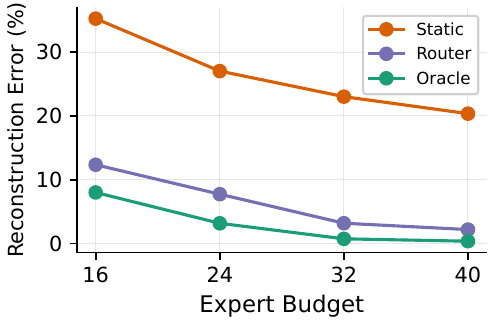}
  \caption{Reconstruction error by selection method on OLMoE-1B-7B, averaged across all MoE layers and 5 benchmarks. Oracle matches Router error using 25\% fewer experts.}
  \label{fig:reconstruction-error}
\end{figure}

Oracle achieves the same reconstruction error as Router while using 25\% fewer experts (\Cref{fig:reconstruction-error}). At $B=24$, Oracle matches Router at $B=32$ (both 3.2\% error); at $B=16$, Oracle nearly matches Router at $B=24$ (8.0\% vs.\ 7.7\%). Static selection incurs 20--35\% error regardless of budget. This 25\% reduction in required experts explains why Oracle maintains quality at lower budgets in \Cref{fig:ablation-selection}.

The gap stems from how each method allocates budget. Router-based selection scores each expert independently by summing routing probabilities across the draft tree. This ignores interactions: if two experts frequently co-activate (e.g., appear together across a draft tree), their contributions to the output are correlated, and selecting both yields diminishing marginal returns. Oracle selection, by contrast, picks experts greedily to minimize reconstruction error. After selecting an expert, the next addition is whichever expert most reduces the residual error. Experts correlated with those already selected provide less marginal benefit and are passed over.

Our profiling shows that routing is concentrated: certain expert pairs co-activate 10--30$\times$ more often than uniform-random selection would predict (averaging 23$\times$ across layers; see \Cref{appendix:coactivation} for details). This concentration is why Oracle matches Router quality with 25\% fewer experts. Router spends budget on popular experts with correlated contributions; Oracle does not. This suggests that selection methods modeling expert co-activation patterns, rather than treating experts independently, could close the gap without requiring full model execution.

\subsection{Expert Coverage Across Datasets}
\label{appendix:coverage}

\begin{figure}
  \centering
  \includegraphics[width=\columnwidth]{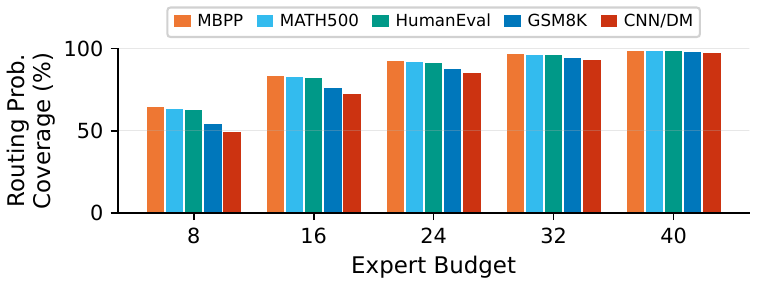}
  \caption{Routing probability coverage by expert budget across datasets on OLMoE-1B-7B with tree size 63. Datasets are ordered by coverage (best to worst): MBPP reaches 95\% coverage with 8 experts, while CNN/DM requires 24 experts to reach 90\%.}
  \label{fig:expert-coverage}
\end{figure}

We profile expert routing on each dataset by running the target model on calibration samples and recording routing probabilities for draft trees of size 63. \Cref{fig:expert-coverage} shows cumulative routing probability captured by the top-$B$ experts.

Routing concentration varies across tasks. Code generation (MBPP, HumanEval) concentrates on few experts: the top 8 capture over 90\% of routing probability. Summarization (CNN/DM) spreads routing more broadly, requiring 24 experts to reach 90\%.

Static ranking uses a fixed expert ordering from calibration data. When routing is concentrated, the same experts dominate across inputs, so a fixed ordering works well. When routing is dispersed, different inputs select different experts, and a fixed ordering misses task-specific patterns. Router-based selection sidesteps this by aggregating probabilities per draft tree.

\section{Related Work}
\label{sec:related}

\paragraph{Speculative Decoding.}
Speculative decoding accelerates autoregressive generation by drafting candidate tokens with a lightweight model and verifying them in parallel with the target model~\citep{speculative-decoding, chen2023accelerating}. Subsequent work improved draft quality through learned drafters: EAGLE~\citep{li2024eagle} predicts future hidden states autoregressively, EAGLE-2~\citep{li2024eagle2} introduces dynamic draft trees based on confidence, and EAGLE-3~\citep{li2025eagle3} scales performance through training-time test. Medusa~\citep{cai2024medusa} adds multiple decoding heads for parallel token prediction. These methods assume verification cost is constant, which holds for dense models but breaks down for MoEs.

\paragraph{Mixture-of-Experts.}
The sparse MoE architecture routes each token to a subset of expert networks, reducing compute while maintaining model capacity~\citep{shazeer2017moe}. Switch Transformers~\citep{fedus2022switch} established the modern paradigm of top-$k$ routing in Transformers, which subsequent architectures including Mixtral~\citep{jiang2024mixtral} and OLMoE~\citep{muennighoff2024olmoe} adopted. On the systems side, MegaBlocks~\citep{gale2023megablocks} provides efficient GPU kernels for MoE computation by avoiding padding through block-sparse operations. Our work targets a different bottleneck: the memory bandwidth cost of loading experts during speculative verification.

\paragraph{Dynamic Sparsity.}
Dense LLMs exhibit heavy-tailed activation patterns: a small fraction of neurons accounts for most of the computation's effect on output quality. Deja Vu~\citep{liu2023deja} exploits this by training a predictor to identify contextual sparsity at runtime, skipping MLP and attention head computations unlikely to affect the output. PowerInfer~\citep{song2024powerinfer} partitions neurons into ``hot'' (frequently activated) and ``cold'' sets based on their Zipfian activation distribution, enabling efficient CPU-GPU hybrid inference by keeping hot neurons on GPU. Both methods target single-token inference in dense models, predicting which neurons to skip for each token independently.

MoE-Spec applies the same principle to a different setting: batched verification in sparse models. Rather than predicting sparsity per token, we aggregate routing probabilities across the draft tree and select experts that collectively cover the batch. This aggregation is key: individual tokens may route to diverse experts, but the distribution over the full tree is heavy-tailed (\cref{fig:motivation-zipf}). Unlike Deja Vu and PowerInfer, we require no learned predictor; the router provides importance scores directly.

\paragraph{MoE Compression.}
MoE-Spec's expert budgeting operates at inference time and is orthogonal to methods that reduce model size permanently. MoE-Pruner~\citep{xie2024moe} prunes weights within experts, while other approaches merge experts~\citep{li2024merge} or apply quantization~\citep{kim2023mixture, frantar2023qmoe}. For memory-constrained settings, offloading methods like MoE-Infinity~\citep{xue2024moe} cache experts between CPU and GPU. MoE-Spec applies on top of compressed or offloaded models.

\paragraph{Speculative Decoding for MoE.}
MoE verification cost scales with draft complexity, creating tension between speculation depth and efficiency. \citet{saxena2025utility} address this by adapting speculation length dynamically, disabling speculation when overhead exceeds benefit. However, their approach targets chain-structured drafts with lengths $K \leq 3$, where expert activation remains moderate.

Modern tree-structured methods like EAGLE-3 use draft trees of 60+ tokens. At this scale, OLMoE activates 54 of 64 experts per layer on average; even modest trees saturate expert activation, limiting what adaptive length can achieve. MoE-Spec takes a different approach: rather than backing off speculation when verification is expensive, we reduce verification cost directly by budgeting experts.

In memory-constrained settings, SP-MoE~\citep{chen2025spmoespeculativedecodingprefetching} uses the drafting stage to prefetch experts via cross-model prediction, while SpecMoEOff~\citep{wang2025acceleratingmixtureofexpertsinferencehiding} employs speculative decoding to amortize CPU-GPU transfer costs. These offloading methods target PCIe bottlenecks; MoE-Spec targets HBM bandwidth when experts fit in GPU memory.

\section{Limitations}
\label{sec:limitations}

MoE-Spec introduces an expert budget hyperparameter that must be selected per deployment. At well-chosen budgets, quality matches EAGLE-3 while improving speedup (\Cref{tab:main-results}); very aggressive budgets can degrade quality, and the Pareto curves in \Cref{fig:pareto-grid-t1} show the full tradeoff space to guide budget selection. The router-based ranking requires computing routing probabilities for all experts across the draft tree before verification, adding 2--3\% overhead; for very small draft trees, this overhead may outweigh the savings from reduced expert loading. We evaluate single-request inference on A100 GPUs, but batch inference changes the tradeoffs: when multiple requests share experts, the marginal cost of loading additional experts decreases, so expert budgeting may provide smaller gains in high-throughput batch serving scenarios.

Our experiments use top-$k$ softmax routing, which is standard in OLMoE, Qwen3, and Mixtral; some recent architectures such as DeepSeek-V3 use sigmoid-based routing with auxiliary-loss-free load balancing, and the aggregation strategy in \Cref{eq:router-score} may need adaptation for these routing schemes. We do not adapt the budget per layer; some MoE layers may tolerate tighter budgets than others, and per-layer adaptation could improve the quality-speedup tradeoff. Additionally, router-based selection treats experts independently; methods that model expert co-activation could close the gap with Oracle selection (\Cref{sec:exp:reconstruction}). We leave these directions to future work.

\section{Conclusion}
\label{sec:conclusion}

Expert routing during MoE verification is heavy-tailed: most of the bandwidth goes to experts that contribute little to output quality. MoE-Spec exploits this by enforcing an expert budget, loading only the highest-scoring experts and dropping the long tail. This decouples verification cost from draft tree complexity, yielding 10--30\% throughput gains over EAGLE-3 across three MoE architectures.

The method requires no learned predictor or auxiliary model. The router already provides importance scores; aggregating them across the draft tree identifies which experts to load. Our Oracle analysis suggests that modeling expert co-activation, rather than scoring experts independently, could push budgets lower still. This points to a natural extension: selection methods that exploit the correlation structure we observe in expert routing.

\newpage

\section*{Impact Statement}
This paper presents work whose goal is to advance the field of machine learning. There are many potential societal consequences of our work, none of which we feel must be specifically highlighted here.

\bibliography{references}
\bibliographystyle{icml2026}

\appendix
\section{Hyperparameters}
\label{appendix:hyperparameters}

\Cref{tab:hyperparameters-general} summarizes the general experimental settings. \Cref{tab:hyperparameters-models} provides model-specific configurations. Note that we use conservative budget ranges for $T=1$ experiments (as reported in \Cref{sec:exp:setup}) and explore the fuller ranges shown in the table for $T=0$ experiments.

\paragraph{Draft Tree Size.}
We use a default draft tree size of 63 tokens, following the EAGLE-3 configuration. The tree size sweep (3--511) in \Cref{fig:mechanism} shows that MoE-Spec's speedup grows with tree size, since larger trees activate more unique experts under standard verification. The choice of 63 balances acceptance rate against verification overhead for the baseline.

\paragraph{Expert Budget Selection.}
Budget ranges are model-specific based on the expert configuration. For OLMoE (64 experts, $k$=8), we sweep $B \in \{8, 16, 24, 32, 40, 48, 56\}$; the minimum of 8 ensures at least one full set of active experts. For Qwen3 (128 experts, $k$=8), we use $B \in \{16, 24, 32, 48, 64\}$. Mixtral's smaller expert pool (8 experts, $k$=2) limits the range to $B \in \{2, 3, 4, 5, 6, 7\}$. Main results (\Cref{tab:main-results}) report the budget maximizing speedup while maintaining quality within one standard deviation of EAGLE.

\paragraph{Reproducibility.}
For $T$=1 experiments, we report mean and standard deviation over 5 random seeds. All models use FP16 precision. We use PyTorch 2.1 with CUDA 12.1. Model weights are loaded from HuggingFace Hub; draft model adapters are from the official EAGLE repository.

\begin{table}
\centering
\caption{General experimental settings.}
\label{tab:hyperparameters-general}
\small
\begin{tabular}{lc}
\toprule
\textbf{Parameter} & \textbf{Value} \\
\midrule
Draft tree size (main) & 63 \\
Draft tree size (sweep) & 3--511 \\
Decoding temperature & 0.0, 1.0 \\
Samples per benchmark & 80 \\
Max generation length & 1024 \\
Shortlist method & Router \\
Token policy & Substitution \\
\midrule
GPU type & NVIDIA A100 \\
GPU memory & 80GB \\
\bottomrule
\end{tabular}
\end{table}

\begin{table}
\centering
\caption{Model-specific configurations.}
\label{tab:hyperparameters-models}
\small
\begin{tabular}{lccc}
\toprule
\textbf{Model} & \textbf{Experts} & \textbf{Budget ($B$)} & \textbf{Draft / GPUs} \\
\midrule
OLMoE-1B-7B & 64 ($k$=8) & 8--56 & EAGLE-3 / 1 \\
Qwen3-30B-A3B & 128 ($k$=8) & 16--64 & EAGLE-3 / 1 \\
Mixtral-8x7B & 8 ($k$=2) & 2--7 & EAGLE / 2 \\
\bottomrule
\end{tabular}
\end{table}

\section{MoE-Spec Algorithm}
\label{appendix:algorithm}

\Cref{alg:moe-spec} provides pseudocode for verification-time expert budgeting. The algorithm has three stages: (1) compute routing probabilities for all tokens in the draft tree, (2) aggregate importance scores per expert by summing routing probabilities, and (3) apply the token policy (truncation or substitution) using only the top-$B$ experts. The aggregation step requires no additional model execution beyond what the router already computes.

\begin{algorithm}
\caption{MoE-Spec Expert Budgeting}
\label{alg:moe-spec}
\begin{tabular}{@{}l@{}}
\textbf{Input:} Hidden states $\{\mathbf{h}_1, \ldots, \mathbf{h}_M\}$, budget $B$, policy \\
\textbf{Output:} MoE output for each token \\[0.5em]
\textit{// Compute routing probabilities} \\
\textbf{for} $t = 1, \ldots, M$ \textbf{do} \\
\quad $g_i(\mathbf{h}_t) \gets \mathrm{softmax}(\mathbf{W}_{\mathrm{gate}} \mathbf{h}_t)$ for all $i$ \\[0.5em]
\textit{// Aggregate importance scores} \\
\textbf{for} $i = 1, \ldots, N$ \textbf{do} \\
\quad $s_i \gets \sum_{t=1}^{M} g_i(\mathbf{h}_t)$ \\[0.5em]
\textit{// Select shortlist} \\
$\mathcal{S} \gets$ top-$B$ experts by $s_i$ \\[0.5em]
\textit{// Apply token policy} \\
\textbf{for} $t = 1, \ldots, M$ \textbf{do} \\
\quad \textbf{if} policy = truncation \textbf{then} \\
\quad\quad $\mathcal{T}_k(\mathbf{h}_t) \gets$ top-$k$ experts by $g_i(\mathbf{h}_t)$ \\
\quad\quad $\mathrm{out}_t \gets \sum_{i \in \mathcal{T}_k(\mathbf{h}_t) \cap \mathcal{S}} \bar{g}_i(\mathbf{h}_t) \cdot E_i(\mathbf{h}_t)$ \\
\quad \textbf{else} \quad \textit{// substitution} \\
\quad\quad $\mathcal{T}_k^{\mathcal{S}}(\mathbf{h}_t) \gets$ top-$k$ in $\mathcal{S}$ by $g_i(\mathbf{h}_t)$ \\
\quad\quad $\mathrm{out}_t \gets \sum_{i \in \mathcal{T}_k^{\mathcal{S}}(\mathbf{h}_t)} \bar{g}_i^{\mathcal{S}}(\mathbf{h}_t) \cdot E_i(\mathbf{h}_t)$ \\[0.5em]
\textbf{return} $\{\mathrm{out}_1, \ldots, \mathrm{out}_M\}$
\end{tabular}
\end{algorithm}

\section{Batched MoE Computation}
\label{appendix:batched}

We use the open-source EAGLE-3 implementation without modification to tree construction or verification logic. Our only change is to the MoE computation: the reference EAGLE implementation processes experts sequentially, gathering tokens routed to each expert, computing the expert output, and accumulating results via \texttt{index\_add\_}. This approach is simple but inefficient, as it issues many small matrix multiplications.

We implement batched expert computation, which processes all active experts in parallel using batched matrix multiplications. Given the set of active experts $\mathcal{E}$, we stack their weight matrices and compute all expert outputs in a single batched \texttt{matmul} operation. This reduces kernel launch overhead and better utilizes GPU parallelism.

\Cref{tab:batched-comparison} compares the two execution modes on OLMoE-1B-7B across three benchmarks. The batched implementation provides 1.4--1.5$\times$ higher throughput with negligible difference in task quality (within run-to-run variance), confirming our optimization does not affect model accuracy. We apply batched execution uniformly to all methods, including AR and EAGLE-3 baselines, to ensure fair comparison.

\begin{table}[t]
\centering
\caption{Comparison of sequential vs.\ batched MoE computation on OLMoE-1B-7B (EAGLE-3, no expert budget). Quality differences are within noise; throughput improves 1.4--1.5$\times$.}
\label{tab:batched-comparison}
\small
\begin{tabular}{llccc}
\toprule
Benchmark & Mode & Quality & Throughput & Accept \\
\midrule
\multirow{2}{*}{GSM8K} & Sequential & 0.663 & 17.1 tok/s & 2.98 \\
 & Batched & 0.688 & 25.2 tok/s & 3.00 \\
\midrule
\multirow{2}{*}{HumanEval} & Sequential & 0.304 & 23.6 tok/s & 4.16 \\
 & Batched & 0.304 & 35.0 tok/s & 4.20 \\
\midrule
\multirow{2}{*}{CNN/DM} & Sequential & 0.180 & 12.3 tok/s & 1.93 \\
 & Batched & 0.173 & 18.4 tok/s & 1.96 \\
\bottomrule
\end{tabular}
\end{table}

\section{Expert Co-activation Analysis}
\label{appendix:coactivation}

To understand why Oracle selection outperforms Router-based selection (\Cref{sec:exp:reconstruction}), we analyze expert co-activation patterns during inference. Co-activation measures how often pairs of experts are selected together for the same token. If certain expert pairs frequently co-activate, their outputs are correlated, and selecting both provides diminishing marginal returns.

\paragraph{Methodology.}
We run inference on 50 samples from each benchmark (GSM8K, HumanEval, MATH500, MBPP, CNN/DM) using OLMoE-1B-7B with EAGLE-3 speculative decoding. For each token processed during verification, we record which experts appear in its top-$k$ selection. For each MoE layer, we build a $64 \times 64$ co-activation matrix where entry $(i, j)$ counts how many times experts $i$ and $j$ were both selected for the same token.

\paragraph{Concentration Metric.}
We measure routing concentration as the ratio between the most frequent co-activating pair and the expected count under uniform-random selection. Under uniform selection of $k=8$ experts from $N=64$, the probability that any specific pair $(i, j)$ both appear is $\frac{k(k-1)}{N(N-1)} \approx 0.014$. If all pairs were equally likely, each would have the same expected count. The concentration ratio is:
\begin{equation}
\text{Concentration} = \frac{\max_{i,j} \text{coact}(i,j)}{\text{expected per pair}}
\end{equation}

\begin{figure}[t]
  \centering
  \includegraphics[width=\columnwidth]{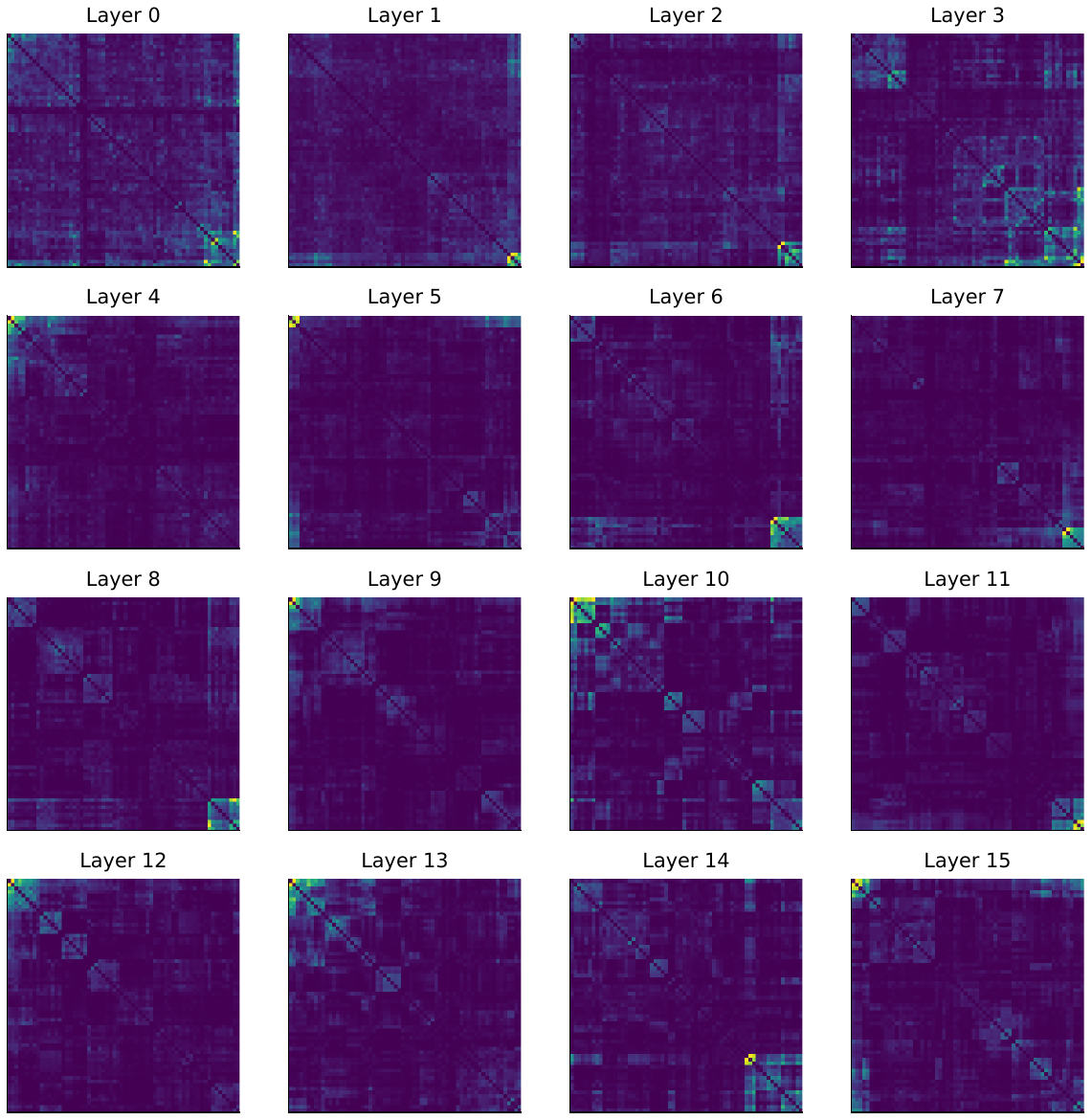}
  \caption{Expert co-activation heatmaps for all 16 MoE layers in OLMoE-1B-7B. Each cell shows normalized pairwise co-activation frequency. Experts are reordered by hierarchical clustering. Bright diagonal blocks indicate expert groups that frequently activate together, with concentration ratios of 10--33$\times$ above uniform-random expectation.}
  \label{fig:coactivation}
\end{figure}

\paragraph{Results.}
\Cref{fig:coactivation} shows co-activation heatmaps for all 16 MoE layers. Experts are reordered using hierarchical clustering to reveal structure. Across layers, certain expert groups co-activate far more frequently than random: concentration ratios range from 10$\times$ to 33$\times$ (mean 23$\times$). The bright diagonal blocks indicate expert clusters that frequently appear together.

This concentration explains the Oracle-Router gap. Router-based selection scores experts independently, so it tends to select multiple experts from the same high-frequency cluster. Oracle selection, which greedily minimizes reconstruction error, avoids this redundancy: after selecting one expert from a cluster, remaining cluster members provide less marginal error reduction and are passed over in favor of experts covering different parts of the output space.

\section{Quality-Speedup Tradeoffs by Benchmark}
\label{appendix:pareto}

\begin{figure}[t!]
  \includegraphics[width=\columnwidth]{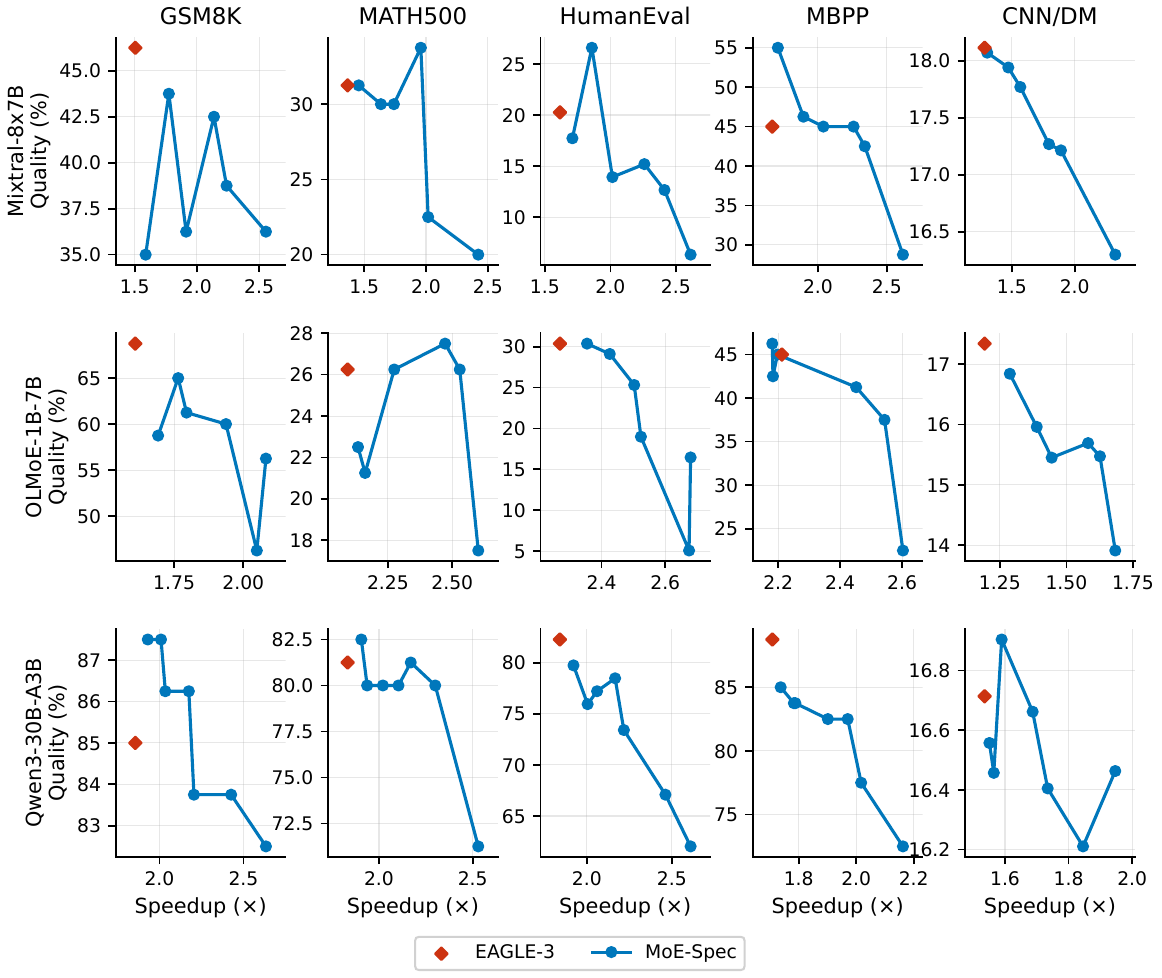}
  \caption{Quality vs.\ speedup at $T=0$ across three models (rows) and five benchmarks (columns). Diamonds mark EAGLE; connected points show MoE-Spec at varying expert budgets.}
  \label{fig:pareto-grid-t0}
\end{figure}

\begin{figure}[t!]
  \centering
  \includegraphics[width=\columnwidth]{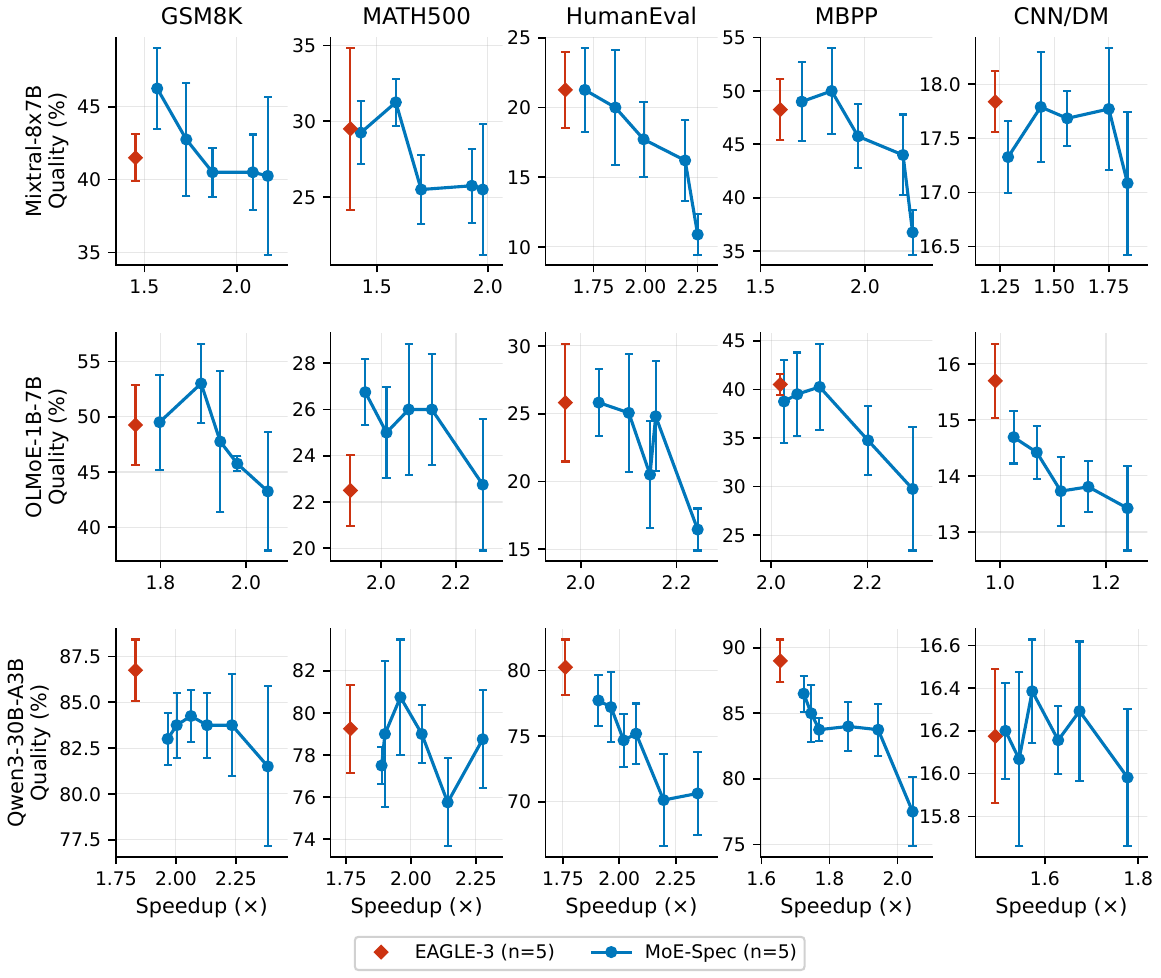}
  \caption{Quality vs.\ speedup at $T=1$ across three models (rows) and five benchmarks (columns). Error bars show standard deviation over 5 seeds.}
  \label{fig:pareto-grid-t1}
\end{figure}

\Cref{fig:pareto-grid-t0,fig:pareto-grid-t1} show the full quality-speedup tradeoff across all model-benchmark combinations at temperatures $T=0$ and $T=1$. Each panel plots task quality against speedup relative to autoregressive decoding. Diamonds mark EAGLE-3 (or EAGLE-1 for Mixtral); connected points trace MoE-Spec across expert budgets.

At $T=0$ (greedy decoding), there is no sampling variance, so each point represents a single deterministic evaluation. At $T=1$, error bars show standard deviation over 5 seeds. In both settings, MoE-Spec extends the Pareto frontier to higher speedups, with quality degradation only at aggressive budgets.

Generally, we find that code generation (HumanEval, MBPP) tolerates tighter budgets than reasoning (GSM8K, MATH500). This matches the coverage analysis in \Cref{appendix:coverage}: code tasks concentrate routing on fewer experts, so budgeting removes less relevant capacity. Summarization (CNN/DM) falls between: routing is dispersed but the task is less sensitive to small output perturbations.

\end{document}